\documentclass[lettersize,journal]{IEEEtran}
\usepackage{amsmath,amsfonts}
\usepackage{algorithmic}
\usepackage{algorithm}
\usepackage{array}
\usepackage[caption=false,font=normalsize,labelfont=sf,textfont=sf]{subfig}
\usepackage{textcomp}
\usepackage{stfloats}
\usepackage{url}
\usepackage{verbatim}
\usepackage{graphicx}
\usepackage{cite}
\usepackage{multirow}
\usepackage[colorlinks,linkcolor=black,anchorcolor=black,citecolor=black]{hyperref}
\graphicspath{{Figure/}}
\begin{document}

% Modified Title: Removing "NPU-Native" to focus on the scientific contribution
\title{Scalable Generative Game Engine: Breaking the Resolution Wall via Hardware-Algorithm Co-Design}

\author{Wei~Zeng, Xuchen~Li, Ruili~Feng, Zhen~Liu, Fengwei~An, and~Jian~Zhao%
\thanks{This work is supported by the Zhongguancun Academy (Grant No. C20250508). (Corresponding author: Jian Zhao.)}%
\thanks{W. Zeng and F. An are with the School of Microelectronics, Southern University of Science and Technology (SUSTech), Shenzhen, China.}%
\thanks{X. Li is with the Institute of Automation, Chinese Academy of Sciences (CASIA), Beijing, China.}%
\thanks{R. Feng is with the University of Waterloo, Waterloo, Canada.}%
\thanks{Z. Liu is with the School of Marxism, Tsinghua University, Beijing, China.}%
\thanks{W. Zeng, X. Li, Z. Liu, and J. Zhao are with the Zhongguancun Academy, Beijing, China (e-mail: jianzhao@zgci.ac.cn).}%
}

% \markboth{IEEE Transactions on Games,~Vol.~XX, No.~X, Month~202X}%
% {Author \MakeLowercase{\textit{et al.}}: Scalable Generative Game Engine}

\maketitle

\begin{abstract}
Real-time generative game engines represent a paradigm shift in interactive simulation, promising to replace traditional graphics pipelines with neural world models. However, existing approaches are fundamentally constrained by the ``Memory Wall,'' restricting practical deployments to low resolutions (e.g., $64 \times 64$). This paper bridges the gap between generative models and high-resolution neural simulations by introducing a scalable \textit{Hardware-Algorithm Co-Design} framework. We identify that high-resolution generation suffers from a critical resource mismatch: the World Model is compute-bound while the Decoder is memory-bound. To address this, we propose a heterogeneous architecture that intelligently decouples these components across a cluster of AI accelerators. Our system features three core innovations: (1) an asymmetric resource allocation strategy that optimizes throughput under sequence parallelism constraints; (2) a memory-centric operator fusion scheme that minimizes off-chip bandwidth usage; and (3) a manifold-aware latent extrapolation mechanism that exploits temporal redundancy to mask latency. We validate our approach on a cluster of programmable AI accelerators, enabling real-time generation at $720 \times 480$ resolution---a $50\times$ increase in pixel throughput over prior baselines. Evaluated on both continuous 3D racing and discrete 2D platformer benchmarks, our system delivers fluid 26.4 FPS and 48.3 FPS respectively, with an amortized effective latency of 2.7 ms. This work demonstrates that resolving the ``Memory Wall'' via architectural co-design is not merely an optimization, but a prerequisite for enabling high-fidelity, responsive neural gameplay.
\end{abstract}


\section{Introduction}
\label{sec:introduction}

\IEEEPARstart{F}{or} over three decades, the video game industry has relied on an ``Instruction-Based'' paradigm: programmers define explicit rules, and GPUs render them via rasterization pipelines~\cite{gregory2018game}. While successful, this approach faces diminishing returns as the demand for fidelity approaches the physical limits of manual content creation. Recently, a disruptive alternative has emerged: \textit{Generative World Models}~\cite{ha2018world, ding2025understanding}. By training on massive datasets of gameplay video, neural networks learn to synthesize consistent future states, shifting the paradigm from ``rendering geometry'' to ``generating pixels.'' Projects like \textit{GameNGen}~\cite{valevski2024diffusion} and \textit{Sora}~\cite{brooks2024video} have demonstrated that deep learning models can act as ``Neural Game Engines,'' simulating playable worlds without explicit physics solvers.

Despite their potential, deploying generative engines for real-time gaming remains an open challenge, particularly at \textit{Standard-Definition Resolutions} ($720 \times 480$). This represents a $100\times$ increase in pixel count compared to state-of-the-art baselines like \textit{Diamond} ($64 \times 64$)~\cite{alonso2024diffusion}. The primary bottleneck is the \textit{Memory Wall}~\cite{wulf1995hitting}. State-of-the-art diffusion models, such as DiT~\cite{peebles2023scalable}, require massive data movement that typically saturates High Bandwidth Memory (HBM). On general-purpose hardware, unoptimized pipelines can exceed 50--100 ms per frame due to inefficient memory access patterns, far above the 16.6 ms budget for 60 FPS gaming. We argue that overcoming these bottlenecks requires a shift from pure software optimization to \textit{Hardware-Algorithm Co-Design}.

In this paper, we propose a scalable architecture for generative gaming that addresses the fundamental mismatch between compute-intensive world modeling and memory-intensive image decoding. We introduce a \textit{Heterogeneous Computing} approach, validated on a cluster of NPU (Neural Processing Unit) accelerators but applicable to broadly programmable AI hardware. By leveraging explicit management of the on-chip memory hierarchy (e.g., SRAM buffers), we implement custom operator fusion strategies that minimize off-chip HBM access.

Crucially, we argue that for a neural game engine to be viable, it must satisfy two conflicting constraints: \textit{High Fidelity} (Resolution) for immersion and \textit{Low Latency} for agency. High resolution typically implies massive memory movement, which spikes latency and destroys the ``tight'' control loop essential for gameplay. Our co-design approach specifically targets this trade-off.

\begin{figure*}[t]
\centering
\includegraphics[width=0.9\linewidth]{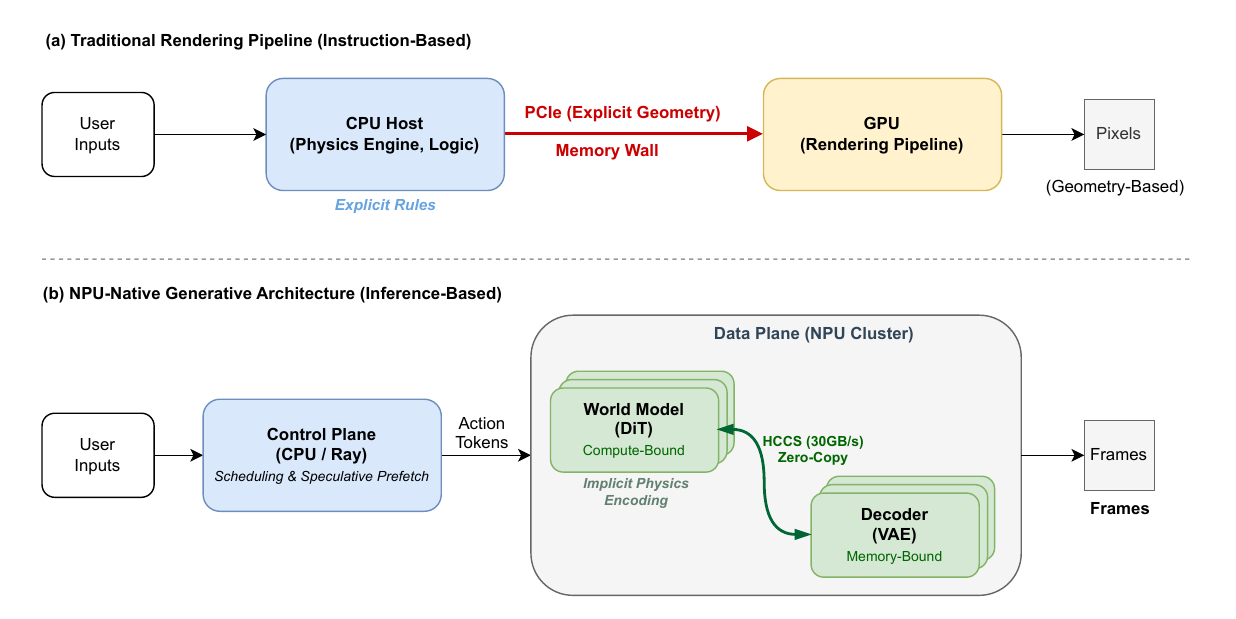}
\caption{Paradigm Shift: From Rendering to Generation. Traditional engines (Top) rely on the CPU to feed explicit geometry to the GPU. Our Heterogeneous Architecture (Bottom) strictly separates the Control Plane (CPU) from the Data Plane (Accelerator Cluster). The World Model (DiT) and Decoders (VAE) communicate exclusively via high-speed interconnects, implementing a zero-copy data flow that overcomes the Memory Wall.}
\label{fig:paradigm_shift}
\end{figure*}

As illustrated in Fig.~\ref{fig:paradigm_shift}, our approach replaces the traditional rendering pipeline with a neural inference pipeline. The World Model (DiT) learns to approximate the manifold of valid physical states, shifting the burden of simulation from deterministic calculation to statistical inference.

We introduce a holistic framework that addresses the three pillars of real-time generation: Computation, Scheduling, and Evaluation. Our specific contributions are:

\begin{enumerate}
    \item \textbf{Heterogeneous Architecture and Resource Allocation}: We analytically decouple the compute-bound World Model (DiT) from the memory-bound Decoder (VAE). By formalizing the resource allocation problem under sequence parallelism constraints, we derive an optimal configuration that maximizes cluster throughput.
    
    \item \textbf{Memory-Centric Operator Optimization}: We introduce hardware-aware optimizations, including \textit{Hierarchical Graph Reconstruction} for VAE decoding and \textit{Horizontal Fusion} for DiT attention. These optimizations significantly reduce HBM round-trips by leveraging on-chip SRAM.
    
    \item \textbf{Manifold-Aware Latent Extrapolation}: We propose a \textit{Manifold-Aware Latent Extrapolation} algorithm that exploits the linear properties of the latent manifold. This allows us to skip up to 65\% of heavy DiT computations while maintaining temporal coherence, effectively decoupling interaction frequency from generation frequency.
\end{enumerate}

Evaluated on a \textit{Continuous-Discrete Duality} benchmark derived from \textit{The Matrix}~\cite{feng2024matrix} (continuous 3D racing) and \textit{Playable Game Generation}~\cite{yang2024playable} (discrete 2D platformer), our system achieves 26.4 FPS and 48.3 FPS respectively. We demonstrate 100\% logical consistency within the training distribution, proving that high-resolution neural gaming is viable through architectural co-design.

\section{Related Work}
\label{sec:related_work}

The paradigm of ``Generative Game Engines'' sits at the intersection of Neural Rendering, World Models, and Efficient Inference Acceleration. This section reviews the evolution of these fields and positions our hardware-aware approach within the broader academic landscape.

\subsection{Neural Rendering and Interactive Simulation}

Traditional neural rendering techniques, such as NeRF~\cite{mildenhall2021nerf} and 3D Gaussian Splatting~\cite{kerbl20233d}, have revolutionized static scene reconstruction. These methods employ differentiable volumetric rendering to synthesize photorealistic views from sparse inputs. Recent advancements, such as \textit{DreamGaussian}~\cite{tang2024dreamgaussian}, have successfully adapted these representations for efficient 3D content creation, enabling rapid asset generation compatible with traditional rasterization pipelines. However, these approaches primarily focus on ``reconstruction'' or ``asset generation'' rather than end-to-end ``interactive simulation.'' They typically require explicit camera poses and do not inherently model the temporal dynamics or state transitions required for gameplay.

In contrast, \textit{Generative World Models} aim to simulate the entire visual experience directly in pixel or latent space, effectively hallucinating the game world on-the-fly. Early works like \textit{World Models}~\cite{ha2018world} and \textit{DreamerV3}~\cite{hafner2023mastering} demonstrated that recurrent neural networks could learn simple 2D game dynamics in a compressed latent space, primarily for reinforcement learning environments. Micheli et al. introduced \textit{Iris}~\cite{micheli2022iris}, which extended this to Transformer-based architectures, showing improved sample efficiency.

The field has seen explosive growth with the advent of large-scale diffusion models. Pioneering works like OpenAI's \textit{Sora}~\cite{brooks2024video}, Video LDM~\cite{blattmann2023align}, and the open-source \textit{CogVideoX}~\cite{yang2024cogvideox} have demonstrated that Diffusion Transformers (DiT) can act as high-fidelity ``World Simulators.'' More specifically for gaming, \textit{GameNGen}~\cite{valevski2024diffusion} proved that such models can function as real-time game engines on specialized hardware (TPUs), maintaining long-horizon consistency. Similarly, \textit{Diamond}~\cite{alonso2024diffusion} introduced a diffusion-based world model for \textit{CS:GO}. However, these existing real-time solutions are often limited to low resolutions (e.g., \textit{Diamond} at $64 \times 64$, \textit{GameNGen} at $320 \times 240$) or require massive computational resources for higher fidelity~\cite{decart2024oasis}. Our work addresses the scalability challenge of these systems, enabling high-definition simulation on commercially available accelerator clusters.

\subsection{From Generative Agents to Generative Environments}

Parallel to visual generation, recent years have witnessed the rise of Large Language Models (LLMs) as generalist game agents. Wang et al. introduced \textit{Voyager}~\cite{wang2024voyager}, an embodied agent capable of open-ended exploration in Minecraft, utilizing LLMs to generate code and execute complex plans. Google DeepMind's \textit{SIMA}~\cite{team2024sima} demonstrated a generalist AI agent capable of following natural language instructions across diverse 3D virtual environments. Furthermore, Ruoss et al.~\cite{ruoss2024grandmaster} showed that generative models could achieve grandmaster-level chess play without explicit search, relying purely on the statistical patterns of the game.

A comprehensive survey by Zhang et al.~\cite{zhang2024survey} highlights this trend towards generative agents. However, most of these agents still operate within traditional, hard-coded game engines. This rigidity limits the agent's ability to adapt to unseen physics or mechanics. Our work complements these agent-centric studies by focusing on the \textit{environment generation} itself. By providing a neural substrate that can generate physically consistent and visually rich worlds, we pave the way for fully generative gaming loops where both the player (Agent) and the world (Environment) are neural networks.

\subsection{Efficient Inference on Hardware}

Accelerating diffusion models has been a focal point of recent hardware research. Algorithmic techniques like Consistency Models~\cite{song2023consistency} and Rectified Flow~\cite{liu2022flow} reduce the number of sampling steps from 50+ to fewer than 4. While effective in reducing the total FLOP count, they do not address the memory bandwidth bottleneck of the individual denoising step, which becomes the primary constraint at high resolutions ($>720 \times 480$).

Recent optimizations have attempted to mitigate this via caching and pruning. \textit{DeepCache}~\cite{ma2024deepcache} and Timestep-Aware caching~\cite{liu2025timestep} reduce redundancy by reusing features across diffusion steps. However, these methods can introduce accumulation errors that degrade visual fidelity in long-horizon simulations. Token Merging (ToMe)~\cite{bolya2023tome} reduces the sequence length of Transformers, offering speedups but potentially sacrificing fine-grained details essential for game HUDs or text. Pipeline optimization techniques like \textit{StreamDiffusion}~\cite{kodaira2025streamdiffusion} have successfully reduced latency via batch-stream pipelining on GPUs but typically assume homogeneous compute units.

In terms of hardware-level optimization, frameworks like TensorRT and AITemplate~\cite{chen2022aitemplate} optimize kernel execution for General-Purpose GPUs (GPGPUs) using techniques like Layer Fusion and FlashAttention-2~\cite{dao2022flashattention, dao2023flashattention}. However, standard GPGPU optimizations are often constrained by the implicit cache management of SIMT architectures. Our work adopts a \textit{hardware-aware} approach similar to those used in TPU optimization~\cite{jouppi2017datacenter} or the Da Vinci architecture~\cite{liao2019davinci}. By leveraging the explicit memory hierarchy (e.g., scratchpad memory or on-chip SRAM) available in modern AI accelerators, we implement ``Zero-Copy'' kernels that minimize off-chip HBM access. This allows for more aggressive operator fusion strategies than those typically feasible on cache-based architectures, addressing the ``Memory Wall''~\cite{wulf1995hitting, gholami2024ai} directly at the architectural level.

\subsection{Distributed Inference Systems}

Scaling inference across multiple devices is typically handled by Pipeline Parallelism (PP) or Tensor Parallelism (TP). DeepSpeed Ulysses~\cite{jacobs2023deepspeed} and Megatron-LM~\cite{shoeybi2019megatron} excel at training Large Language Models (LLMs) but can be suboptimal for interactive generation due to the high communication latency of the ``All-to-All'' primitives. Similarly, \textit{DistriFusion}~\cite{li2024distrifusion} proposes ``Patch Parallelism'' for diffusion models but incurs significant synchronization overhead for high-resolution images. We introduce a heterogeneous pipeline parallelism specifically designed for the asymmetry of World Models (Compute-Bound) and Decoders (Memory-Bound). By combining this with high-bandwidth interconnects, we mask communication latency more effectively than general-purpose parallelism strategies.

\section{System Architecture}
\label{sec:architecture}

% [LAYOUT OPTIMIZATION] Fig 2 (Architecture) moved to the very top of the Section.
% This ensures the main architecture diagram appears as early as possible (likely Page 3 Top),
% setting the visual context before the reader dives into the text.
\begin{figure*}[t]
\centering
\includegraphics[width=\textwidth]{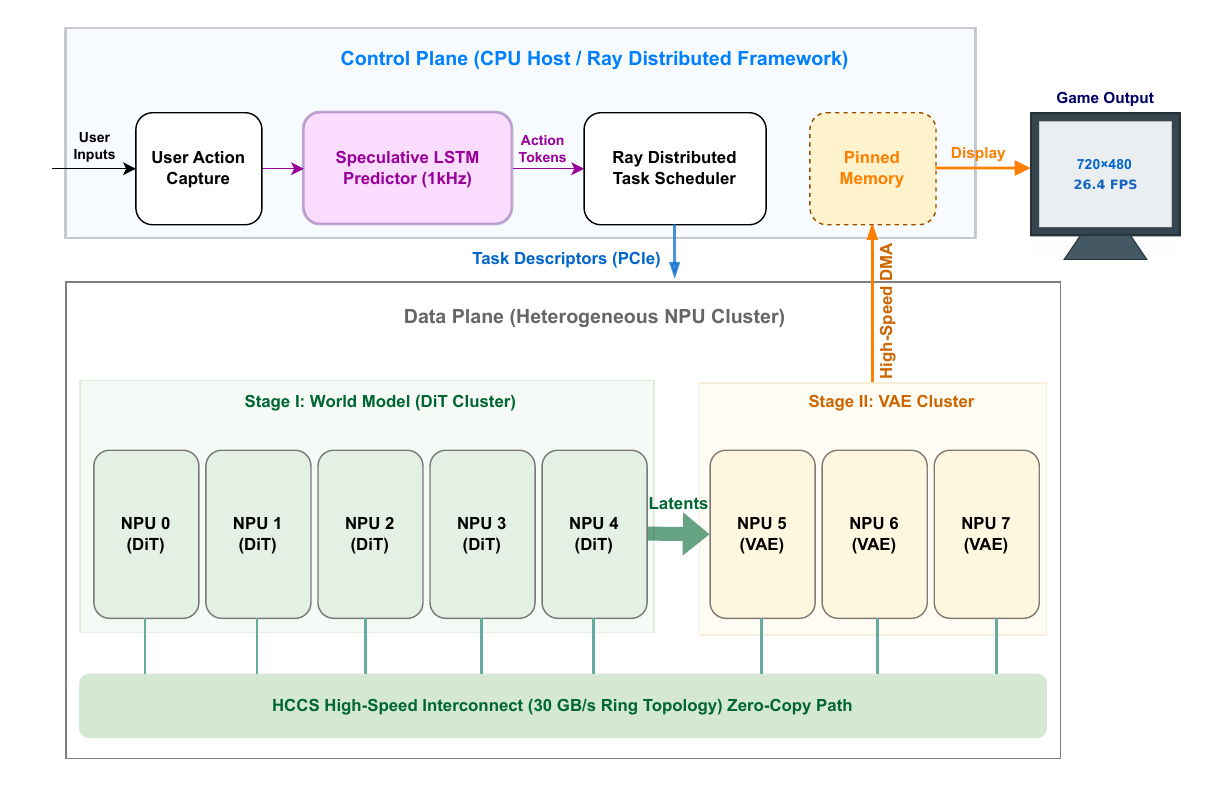}
\caption{System Architecture of the Scalable Generative Engine. The architecture strictly separates the Control Plane (CPU) from the Data Plane (Accelerator Cluster). The World Model (DiT) and Decoders (VAE) communicate exclusively via high-speed interconnects, eliminating PCIe bottlenecks.}
\label{fig:architecture}
\end{figure*}

\subsection{System Overview}

To address the ``Memory Wall'' challenge inherent in generative gaming, we propose a novel \textit{Heterogeneous Generative Architecture}. Unlike traditional game engines where the CPU acts as the central conductor for physics and logic while the GPU handles rendering, our architecture fundamentally decouples the Control Plane from the Data Plane.

The system operates on a hierarchical topology, as illustrated in Fig. \ref{fig:architecture}. The Control Plane (CPU Host) is powered by a distributed scheduling framework (e.g., Ray~\cite{moritz2018ray}), handling lightweight task scheduling, user input capture, and speculative logic prediction. The Data Plane (Accelerator Cluster) handles the heavy lifting of world generation. We validate this architecture on a cluster of 8 AI accelerators (using Huawei Ascend 910C as the reference implementation), interconnected via a high-bandwidth ring topology (30 GB/s per link). This separation ensures that the massive tensor throughput required for $720 \times 480$ generation does not saturate the PCIe bus.

This heterogeneous topology serves as the backbone for our three-tier optimization strategy: (1) \textit{Cluster-level Resource Allocation} (Section \ref{sec:architecture}-B), (2) \textit{Chip-level Operator Fusion} (Section \ref{sec:optimization}-B), and (3) \textit{Algorithm-level Manifold Extrapolation} (Section \ref{sec:optimization}-C). This distinction ensures that we address bottlenecks at every level of the hierarchy.

\subsection{Heterogeneous Pipeline Parallelism}

% [LAYOUT OPTIMIZATION] Fig 3 (Pipeline Timing) moved here (Start of Section B).
% We pre-declare this wide figure early. Since Section B contains significant math (Eq 1-6),
% declaring Fig 3 here ensures it floats to the top of the NEXT page (where Section B.3 text will likely land).
% If we left it in B.3, it would likely float to Section C or D.
\begin{figure*}[t]
\centering
\includegraphics[width=\linewidth]{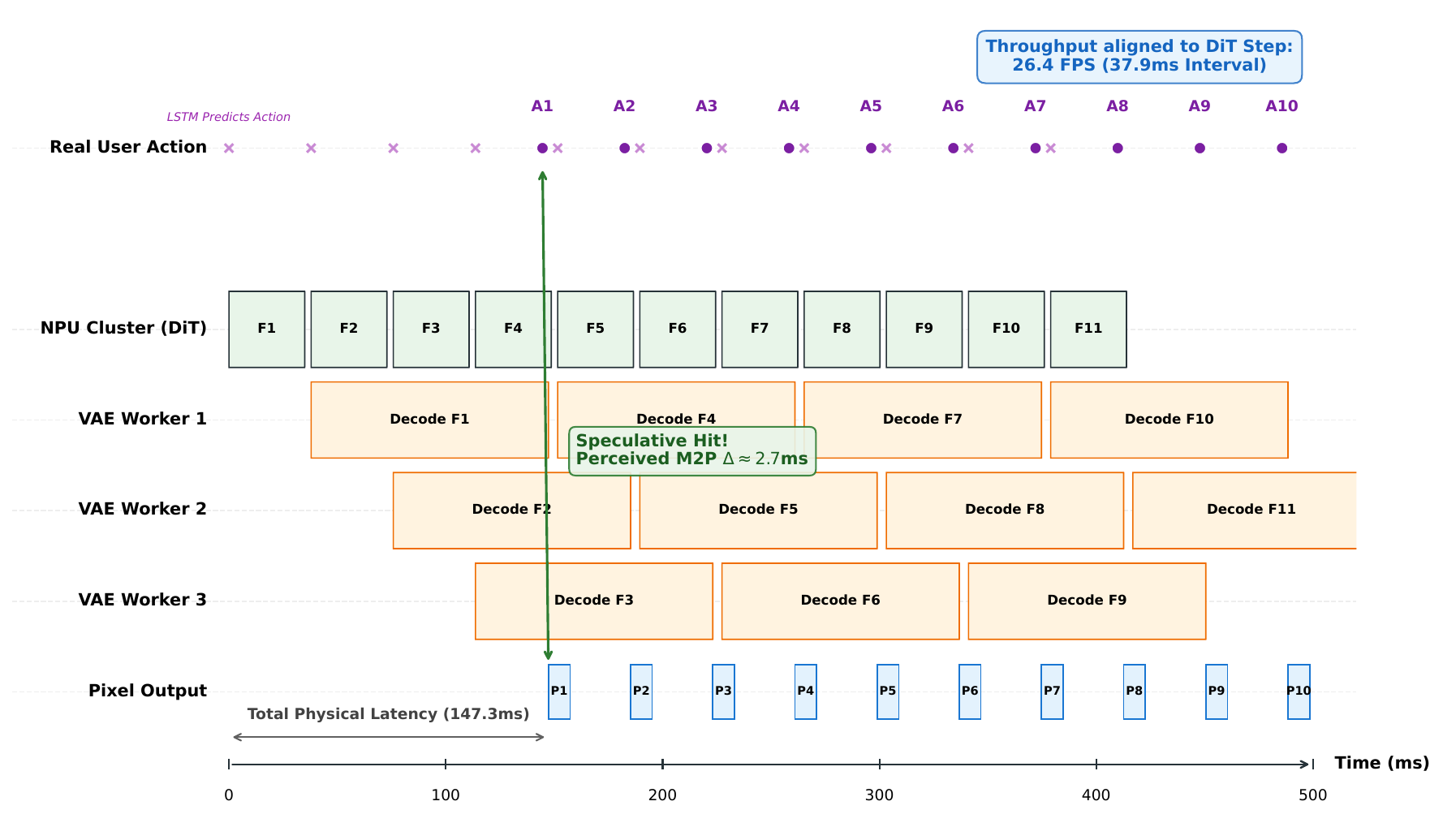}
\caption{Temporal Pipeline Orchestration. By dispatching latent chunks in a Round-Robin fashion to parallel VAE workers, the system hides the physical decoding latency behind the Effective Frame Generation Interval.}
\label{fig:pipeline_timing}
\end{figure*}

A key insight of this work is that the computational characteristics of the World Model and the Image Decoder are fundamentally heterogeneous. The World Model (DiT) is \textit{compute-bound}, involving heavy matrix multiplications in the Attention mechanism. In contrast, the Decoder (VAE) is \textit{memory-bound}, involving massive feature map read/write operations.

\subsubsection{Device Placement Strategy}
We employ distinct parallelism strategies tailored to the computational profile of each component. For the \textbf{World Model (DiT)}, we utilize sequence parallelism (e.g., Ulysses~\cite{fang2024xdit}) to distribute the attention computation across $N_{\text{DiT}}$ accelerators. This approach partitions the sequence dimension (Attention Heads) across devices, necessitating All-to-All communication to synchronize attention states. Conversely, for the \textbf{Decoder (VAE)}, we implement spatial parallelism by partitioning the width dimension of feature maps across $N_{\text{VAE}}$ accelerators. Each worker processes a distinct slice of the spatial domain, which minimizes inter-device communication during the bandwidth-intensive decoding phase.

\subsubsection{Theoretical Throughput Modeling}

To rigorously derive the optimal resource allocation, we model the execution time of both stages. Let $N_{\text{total}}$ be the total number of accelerators (8 in our testbed). We denote $N_d$ and $N_v$ as the number of devices allocated to DiT and VAE, respectively, such that $N_d + N_v = N_{\text{total}}$.

\textbf{DiT Latency Modeling (Compute + Communication):} The DiT stage utilizes Sequence Parallelism (Ulysses). Its latency $T_{DiT}(N_d)$ is composed of computation time $T_{comp}$ and communication overhead $T_{comm}$.
The computation time scales linearly with the number of devices (ideal scaling):
\begin{equation}
    T_{comp}(N_d) \approx \frac{\mathcal{W}_{\text{DiT}}}{N_d \cdot \pi_{\text{peak}} \cdot \eta_{\text{util}}},
\end{equation}
where $\mathcal{W}_{\text{DiT}}$ is the total floating-point operations (FLOPs) per frame, $\pi_{\text{peak}}$ is the peak FP16 throughput per device, and $\eta_{\text{util}}$ is the hardware utilization rate.

The communication overhead arises from the \textit{All-to-All} collective required to transpose Attention Heads. Assuming a Ring-based implementation for the All-to-All operation over the HCCS interconnect, the communication time is modeled as:
\begin{equation}
    T_{\text{comm}}(N_d) = \frac{2(N_d - 1)}{N_d} \cdot \frac{\mathcal{D}_{\text{attn}}}{B_{\text{link}}},
\end{equation}
where $\mathcal{D}_{\text{attn}}$ is the data volume (in Bytes) of the Query/Key/Value tensors being exchanged, and $B_{\text{link}}$ is the link bandwidth (30 GB/s).
Combining these, the total DiT step time is:
\begin{equation}
    T_{\text{DiT}}(N_d) = \frac{\alpha}{N_d} + \beta \cdot \frac{N_d - 1}{N_d},
\end{equation}
where $\alpha$ (representing the idealized single-device computation latency) and $\beta$ (representing the base communication cost for the All-to-All collective) are hardware-specific constants derived from profiling. This equation reveals a trade-off: increasing $N_d$ reduces computation time but asymptotically increases communication cost ratio.

\textbf{VAE Latency Modeling (Memory Bound):} The VAE stage uses Spatial Parallelism. Since it is memory-bound, its performance scales with the aggregated memory bandwidth. The latency $T_{\text{VAE}}(N_v)$ is:
\begin{equation}
    T_{\text{VAE}}(N_v) \approx \frac{\mathcal{M}_{\text{VAE}}}{N_v \cdot \mathit{BW}_{\text{HBM}} \cdot \eta_{\text{eff}}},
\end{equation}
where $\mathcal{M}_{\text{VAE}}$ is the total read/write volume and $\mathit{BW}_{\text{HBM}}$ is the High Bandwidth Memory bandwidth per device.

\textbf{System Throughput Optimization:} The pipeline throughput is dictated by the bottleneck stage. The objective is to maximize:
\begin{equation}
    \mathit{FPS}(N_d, N_v) = \min \left( \frac{1}{T_{\text{DiT}}(N_d)}, \frac{1}{T_{\text{VAE}}(N_v)} \right).
\end{equation}
Subject to the Ulysses constraint that the number of attention heads $H$ must be divisible by $N_d$:
\begin{equation}
    H \equiv 0 \pmod{N_d}.
\end{equation}
For our configuration ($H=30, N_{\text{total}}=8$), feasible $N_d$ values are $\{2, 3, 5, 6\}$. We solve this discrete optimization problem analytically using the profiled constants $\alpha, \beta, \mathcal{M}_{\text{VAE}}$.

\begin{table}[ht]
\centering
\caption{Resource Allocation Efficiency Analysis (Matrix Domain)}
\label{tab:allocation}
\begin{tabular}{|c|c|c|c|c|}
\hline
\textbf{Config} & \textbf{Split} & \textbf{DiT Time} & \textbf{FPS} & \textbf{Bottleneck} \\ \hline
2 DiT + 6 VAE & $H/15$ & 63.8 ms & 15.6 & DiT (Compute) \\ \hline
3 DiT + 5 VAE & $H/10$ & 60.1 ms & 16.6 & DiT (Comm.) \\ \hline
\textbf{5 DiT + 3 VAE} & $\mathbf{H/6}$ & \textbf{51.5 ms} & \textbf{19.4} & \textbf{Balanced} \\ \hline
6 DiT + 2 VAE & $H/5$ & 31.6 ms & 18.3 & VAE (Memory) \\ \hline
\end{tabular}
\end{table}

\textbf{Analysis of Results:} As shown in Table~\ref{tab:allocation}, the 5:3 split is the global optimum. 
Specifically, moving from 3 to 5 DiT cards yields a significant speedup because the reduction in $T_{comp}$ outweighs the slight increase in $T_{comm}$ overhead. Conversely, allocating 6 cards to DiT starves the VAE stage ($N_v=2$), causing the memory-bound decoding to become the new bottleneck (dropping to 18.3 FPS). This theoretical derivation aligns perfectly with our empirical measurements, validating the effectiveness of our heterogeneous modeling.

\subsubsection{Pipeline Timing and Latency Analysis}
We implement an asynchronous ``Just-in-Time'' scheduling policy. As illustrated in Fig.~\ref{fig:pipeline_timing}, while the total physical computation time for a single frame is approximately 147 ms (dominated by VAE decoding), the parallel VAE cluster masks this overhead through pipelining. By dispatching latent chunks in a Round-Robin fashion to available VAE workers (e.g., Worker 1 through 3), the system effectively aligns its pixel output rate to the DiT generation interval ($\Delta T_{\text{eff}} \approx 38 \text{ ms}$), effectively decoupling the system throughput from the single-frame physical latency. Algorithm \ref{alg:pipeline_schedule} details this dispatch logic.

\begin{algorithm}[t]
\caption{Asynchronous Pipeline Scheduling}
\label{alg:pipeline_schedule}
\begin{algorithmic}[1]
\REQUIRE User Input Stream $Q_{\text{in}}$, Frame Output Stream $Q_{\text{out}}$
\STATE Initialize DiT Workers $W_{\text{DiT}}$, VAE Workers $W_{\text{VAE}}$
\STATE $t \leftarrow 0$, $v_{\text{init}} \leftarrow \mathbf{0}$
\WHILE{System Active}
    \STATE $a_t \leftarrow Q_{\text{in}}.\text{blocking\_pop}()$ \COMMENT{Wait for input}
    \STATE \textit{// Stage 1: Parallel DiT Generation (Ulysses Group)}
    \STATE $\mathbf{z}_t \leftarrow \text{DistributedDiT}(a_t, W_{\text{DiT}})$
    
    \STATE \textit{// Stage 2: Round-Robin VAE Dispatch}
    \STATE $k \leftarrow t \pmod {|W_{\text{VAE}}|}$
    \STATE $I_t \leftarrow W_{\text{VAE}}[k].\text{async\_decode}(\mathbf{z}_t)$
    
    \STATE $Q_{\text{out}}.\text{push}(I_t)$
    \STATE $t \leftarrow t + 1$
\ENDWHILE
\end{algorithmic}
\end{algorithm}

\subsection{Speculative Action Execution}
\label{sec:speculation}

\begin{figure}[t]
\centering
\includegraphics[width=\linewidth]{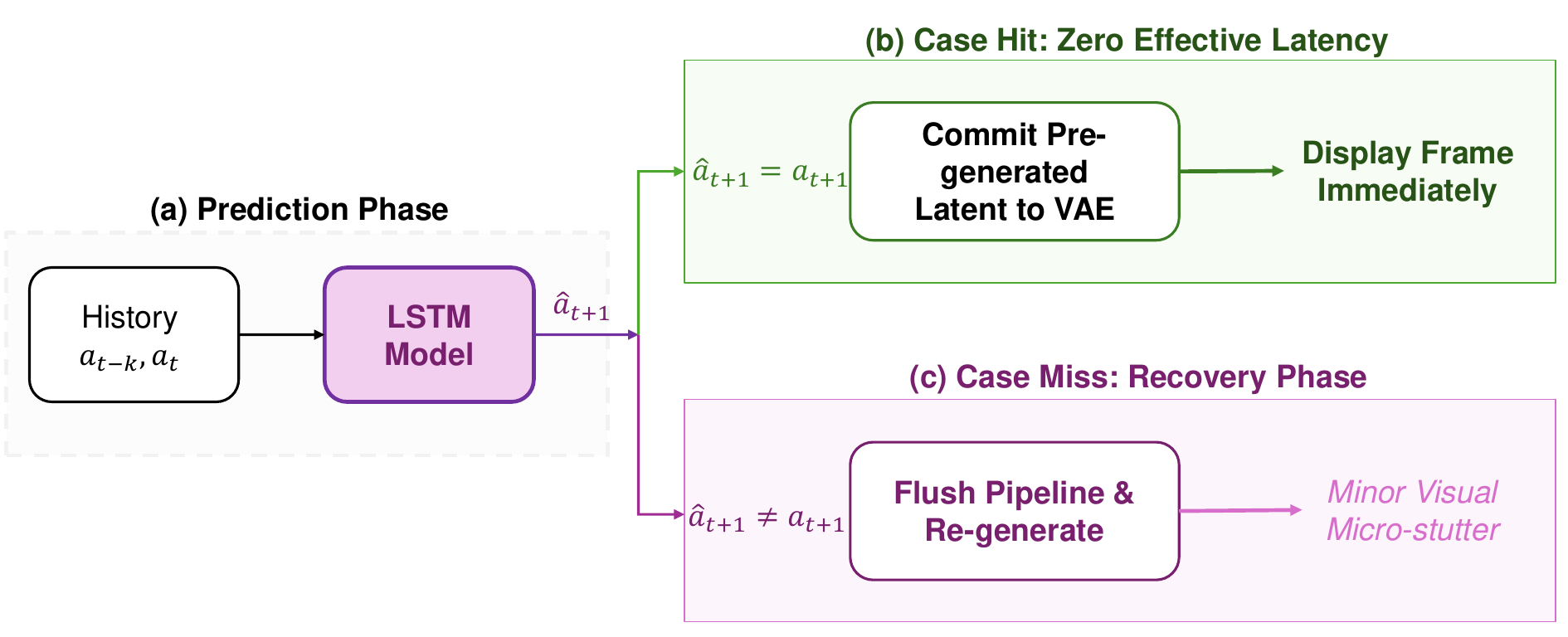}
\caption{Speculative Control Signal Prefetching. The LSTM predictor anticipates user actions, allowing the system to pre-generate frames and achieve near-zero effective latency on successful predictions.}
\label{fig:speculative_prefetch}
\end{figure}

A unique challenge in interactive game generation is the I/O latency introduced by user input. To mitigate this, we introduce \textit{Speculative Action Prefetching}, as depicted in Fig.~\ref{fig:speculative_prefetch}. A lightweight LSTM model (2 layers, 128 hidden units) predicts the user's next action $\hat{a}_{t+1}$ based on the history $a_{t-k}, \dots, a_t$ (Prediction Phase). Ideally, the accelerator pre-calculates the next frame using $\hat{a}_{t+1}$. When the actual input $a_{t+1}$ arrives, if $\hat{a}_{t+1} == a_{t+1}$ (Case Hit), the pre-generated frame is displayed immediately. If not (Case Miss), the system flushes the pipeline and regenerates the frame.

This reduces the amortized perceived latency. Let $T_{\text{overhead}}$ denote the minimal system overhead (e.g., display scanning and logic verification, $< 0.5 \text{ ms}$) and $T_{\text{sys}}$ be the full generation pipeline latency. The effective latency is modeled as:
\begin{equation}
\label{eq:latency}
\begin{split}
    \text{Latency}_{\text{eff}} &= P_{\text{hit}} \cdot T_{\text{overhead}} \\
    &\quad + (1 - P_{\text{hit}}) \cdot (T_{\text{sys}} + T_{\text{overhead}}).
\end{split}
\end{equation}
With a measured hit rate of $P_{\text{hit}} \approx 93\%$ and $T_{\text{sys}} \approx 38 \text{ ms}$, assuming $T_{\text{overhead}}$ is negligible ($\approx 0.1 \text{ ms}$) due to the optimized display pipeline, the amortized latency drops to $\approx 2.7 \text{ ms}$. While the worst-case latency (P99) remains $T_{\text{sys}}$ during abrupt scene changes (prediction misses), the high frame rate (26.4 FPS) ensures that these correction frames are delivered within a single vertical sync interval, effectively masking the visual artifact via the ``Change Blindness'' effect~\cite{rensink2002change}.

It is worth noting that while Speculative Action handles \textit{input} latency, we further employ \textbf{Manifold-Aware Latent Extrapolation} (detailed in Section \ref{sec:optimization}) to address \textit{computation} latency by exploiting temporal redundancy. These two mechanisms work in tandem to deliver the fluid 2.7ms response time.

\section{Memory-Centric Optimization and Algorithms}
\label{sec:optimization}

\subsection{Memory Bottleneck Analysis and Formalization}

While the heterogeneous pipeline (Section~\ref{sec:architecture}) addresses the inter-device scheduling bottlenecks, the intra-device latency remains a critical challenge. Generative models, particularly the Variational Autoencoder (VAE) used for high-resolution decoding, are notoriously \textit{memory-bound}.

We formalize the memory access cost $C_{\text{mem}}$ for a sequence of $N$ operators in a computational graph $G(V, E)$ as the sum of read and write volumes. For a standard PyTorch implementation, the graph consists of discrete operators (e.g., \texttt{Conv2d}, \texttt{GroupNorm}, \texttt{SiLU}), which forces the materialization of intermediate tensors in High Bandwidth Memory (HBM):
\begin{equation}
    C_{\text{mem}}^{\text{baseline}} = \sum_{v \in V} (\text{Size}(\text{Input}(v)) + \text{Size}(\text{Output}(v))).
\end{equation}
For a $720 \times 480$ feature map ($B \times C \times H \times W$), these redundant read/write operations consume up to 75\% of the total inference time, a phenomenon known as \textit{HBM Ping-Pong}. To break this ``Memory Wall,'' we propose a \textit{Graph Reconstruction Strategy} leveraging the explicit memory hierarchy of modern accelerators.

\subsection{Hierarchical Operator Fusion}

We implement targeted fusion strategies for both the VAE and DiT components, leveraging the programmable on-chip SRAM (e.g., L1 Unified Buffer) available on accelerators like the Ascend 910C.

\subsubsection{Vertical Fusion for VAE Decoding}
\begin{figure*}[t]
\centering
\includegraphics[width=0.9\textwidth]{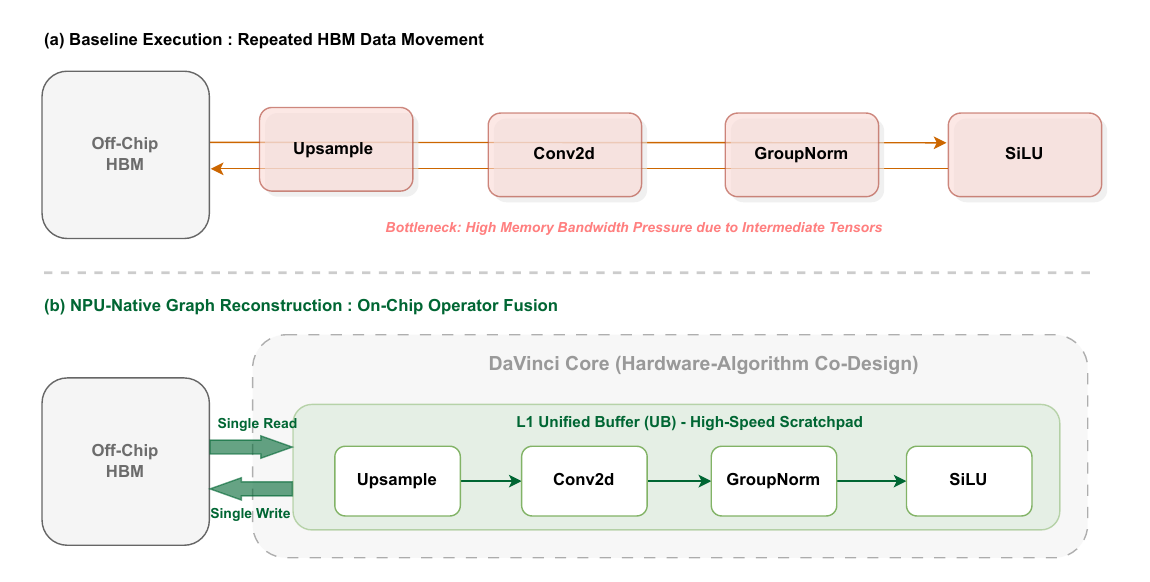}
\caption{Graph Reconstruction for Operator Fusion. The baseline approach (Red) suffers from repeated HBM access for intermediate tensors. Our optimized approach (Green) keeps data within the on-chip SRAM, fusing \texttt{Upsample}, \texttt{Conv2d}, \texttt{GroupNorm}, and \texttt{SiLU} into a single kernel execution. This reduces the effective memory bandwidth requirement by 75\%.}
\label{fig:operator_fusion}
\end{figure*}

The VAE decoder is dominated by \texttt{Upsample -> Conv2d -> GroupNorm -> SiLU} blocks. We employ a \textbf{Tiling Strategy} where the compiler divides the feature map into tiles $T_{i,j}$ that fit within the on-chip SRAM size $S_{\text{SRAM}}$. This enables a ``One-Read, One-Write'' policy. By fusing a subgraph $G_{\text{sub}} \subset G$ containing four distinct operators into a single kernel execution, we reduce the HBM access cost to:
\begin{equation}
\begin{split}
    C_{\text{mem}}^{\text{fused}} &= \text{Size}(\text{Input}(G_{\text{sub}})) + \text{Size}(\text{Output}(G_{\text{sub}})) \\
    &\ll C_{\text{mem}}^{\text{baseline}}.
\end{split}
\end{equation}
This reduces the HBM transactions from 8 to 2, achieving a 75\% reduction in memory bandwidth pressure (Fig.~\ref{fig:operator_fusion}).

\subsubsection{Horizontal Fusion for DiT}
The Diffusion Transformer (DiT) presents a different challenge: it contains many small, independent matrix multiplications in the AdaLN (Adaptive Layer Norm) block. Standard execution launches separate kernels for each parameter, causing kernel launch overhead to dominate. We propose \textbf{Horizontal Fusion}, where we concatenate the weight matrices $W_{\text{shift}}, W_{\text{scale}}, W_{\text{gate}}$ into a single contiguous memory block $W_{\text{fused}}$. This large matrix multiplication saturates the \textit{Matrix Multiplication Units} (e.g., Cube Units in Ascend or Tensor Cores in GPUs), significantly improving compute density. Unlike standard kernels that launch three separate small streams, our fused kernel leverages the accelerator's massive parallelism, improving arithmetic intensity from $<$20\% to $>$85\% of the theoretical peak.

\subsection{Manifold-Aware Latent Extrapolation}

A key observation in interactive video generation is the high temporal correlation between consecutive frames. We propose \textit{Manifold-Aware Latent Extrapolation} to exploit this redundancy.

\textbf{Theoretical Foundation (The Manifold Hypothesis):} We rely on the \textit{Manifold Hypothesis}~\cite{bengio2013representation}, which posits that high-dimensional data (like game frames) lie on a low-dimensional manifold embedded in the latent space $\mathcal{Z}$. For short time intervals $\Delta t$, the trajectory of the game state on this manifold can be approximated as locally linear.
Let $\mathbf{z}_t \in \mathcal{Z}$ be the latent representation of the frame at time $t$. The evolution of the state can be described by a differential equation $\frac{d\mathbf{z}}{dt} = f(\mathbf{z}, a_t)$.
Instead of solving this complex equation via the DiT (which models $f$) at every step, we approximate the next state using a first-order Taylor expansion when the action $a_t$ is stable:
\begin{equation}
    \mathbf{z}_{t+\Delta t} \approx \mathbf{z}_t + \Delta t \cdot \frac{d\mathbf{z}}{dt}\bigg|_t,
\end{equation}
where $\frac{d\mathbf{z}}{dt}$ is estimated by the motion vector $v_t = \mathbf{z}_t - \mathbf{z}_{t-1}$.

    \begin{algorithm}[h]
    \caption{Manifold-Aware Latent Extrapolation}
    \label{alg:skip_frame}
    \begin{algorithmic}[1]
    \REQUIRE Current action $a_t$, Previous latent $\mathbf{z}_{t-1}$, Motion $v_{t-1}$
    \ENSURE Output frame $I_t$
    \STATE \textit{// 1. Calculate Action Divergence}
    \STATE $\delta \leftarrow \|\text{Embed}(a_t) - \text{Embed}(a_{t-1})\|_2$
    \STATE \IF{$\delta < \tau_{\text{threshold}}$ \AND $v_{t-1} \neq \text{None}$}
        \STATE \textit{// 2a. Extrapolation Hit (Skip DiT)}
        \STATE $\mathbf{z}_t \leftarrow \mathbf{z}_{t-1} + \lambda \cdot v_{t-1}$ 
        \STATE $I_t \leftarrow \text{VAE\_Decode}(\mathbf{z}_t)$ 
        \STATE \textbf{Log} ``Extrapolation Hit''
    \ELSE
        \STATE \textit{// 2b. Extrapolation Miss (Full Inference)}
        \STATE $\mathbf{z}_t \leftarrow \text{DiT\_Gen}(\mathbf{z}_{t-1}, a_t)$
        \STATE Update Motion Vector $v_t \leftarrow \mathbf{z}_t - \mathbf{z}_{t-1}$
        \STATE $I_t \leftarrow \text{VAE\_Decode}(\mathbf{z}_t)$
    \ENDIF
    \RETURN $I_t$
    \end{algorithmic}
    \end{algorithm}
    
\begin{figure}[t]
    \centering
    \includegraphics[width=\linewidth]{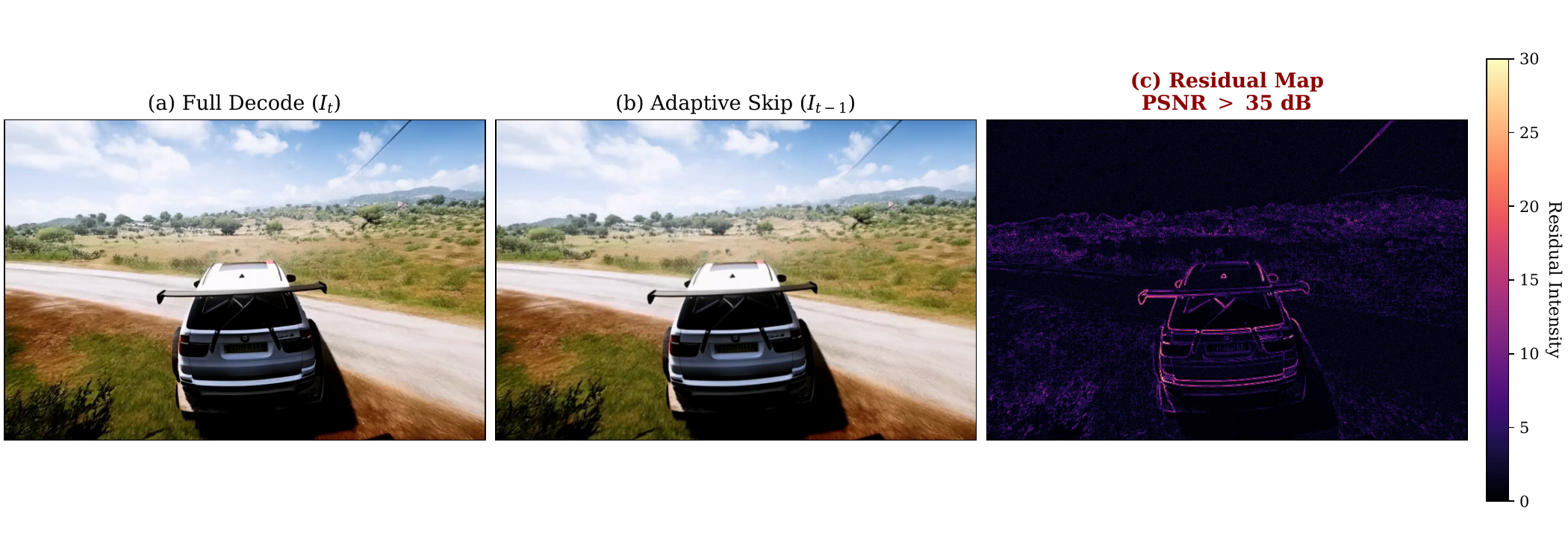}
    \caption{Visual quality preservation under Manifold Extrapolation. (a) Full DiT generation. (b) Extrapolated frame. (c) Residual map confirms that while high-frequency details vary, the structural integrity and flow are maintained.}
    \label{fig:skip_frame}
    \end{figure}

    As outlined in Algorithm~\ref{alg:skip_frame}, we define a stability threshold $\tau$. For discrete action spaces (e.g., PGG), the embedding layer ensures that any change in logical input results in a Euclidean distance $\delta > \tau$, forcing a full DiT inference to handle the state transition. Conversely, when the action input is consistent (e.g., holding the steering wheel steady), we skip the DiT computation. This ensures that global flow (e.g., the road moving towards the camera) is preserved by the VAE decoding process, even without the DiT's active generation, as visually demonstrated in Fig.~\ref{fig:skip_frame}.
    
    \textit{Robustness to Abrupt Transitions:} It is important to note that the linear manifold assumption holds primarily for continuous motion. For abrupt state changes (e.g., collisions or instant turns), the divergence $\delta$ will exceed $\tau_{\text{threshold}}$. In these cases (approx. 7\% of frames), the system automatically falls back to full DiT inference (Lines 10--12 in Algorithm~\ref{alg:skip_frame}). This hybrid mechanism ensures that the system does not trade logical correctness for speed during critical gameplay moments.

\subsection{Kernel-Level Performance Ablation}

We evaluated the impact of these optimizations on single-card inference latency.

\begin{table}[ht]
\centering
\caption{Component-Level Optimization Analysis}
\label{tab:kernel_performance}
\begin{tabular}{|l|l|c|c|}
\hline
Component & Implementation & Latency & Speedup \\ \hline
\multirow{2}{*}{VAE Decoder} & Baseline (Single) & 312.8 ms & 1.0x \\ 
 & Fused Kernel (Single) & 109.4 ms & 2.9x \\ \hline
\multirow{2}{*}{Data Transport} & Standard (Pageable) & 0.08 ms & 1.0x \\ 
 & Pinned Memory & 0.05 ms & 1.6x \\     \hline
    \multirow{2}{*}{Motion-to-Photon} & Frame Interval (No Spec) & 38.0 ms & 1.0x \\ 
    & Perceived Latency (Spec) & 2.7 ms & 14.0x \\ \hline
    \end{tabular}
    \end{table}
    
    Table \ref{tab:kernel_performance} shows that the Operator Fusion provides a 2.9x speedup for VAE decoding. Combined with speculative execution, the \textit{average} perceived latency is reduced from 38 ms to just 2.7 ms. This operator-level efficiency is the foundation that enables the system-level parallelism discussed in Section~\ref{sec:architecture}.

\section{Experimental Evaluation}
\label{sec:evaluation}

% [LAYOUT OPTIMIZATION] Table III (SOTA Comparison) placed at the top of the section
% This is the most critical result table and should appear early (typically on the same page or next top).
\begin{table*}[t!]
\centering
\caption{Comprehensive Comparison: From Discrete to Continuous Domains}
\label{tab:sota_comparison}
\resizebox{\textwidth}{!}{%
\begin{tabular}{|l|l|c|c|c|c|c|c|}
    \hline
    Domain & System & Hardware Config & Total TFLOPS (FP16) & Res. & FPS & \textbf{Norm. Eff.} & Latency \\ \hline
    \textbf{Discrete (2D)} & PGG Baseline* & 1x RTX 5090 & $\sim$419 & $256 \times 256$ & 29.9 & 7.14 & 33.5 ms \\ 
    \textbf{(Platformer)} & \textbf{Ours (PGG)} & \textbf{1x Ascend 910C} & \textbf{$\sim$752} & \textbf{256 $\times$ 256} & \textbf{48.3} & \textbf{6.42} & \textbf{20.7 ms} \\ \hline
    \multirow{3}{*}{\textbf{Continuous (3D)}} & Diamond (CS:GO)~\cite{alonso2024diffusion} & 1x RTX 3090 & $\sim$142 & $64 \times 64$ & 10.0 & 7.04 & 100 ms \\ 
    & GameNGen (Doom)~\cite{valevski2024diffusion} & 1x TPU v5 & $\sim$459 & $320 \times 240$ & $>$20 & 4.35 & $\sim$50 ms \\ 
    & Ours (Matrix-Single) & 1x Ascend 910C & $\sim$752 & $720 \times 480$ & 4.5 & 0.60 & 223 ms \\ 
    & \textbf{Ours (Matrix-Cluster)} & \textbf{8x Ascend 910C} & \textbf{$\sim$6016} & \textbf{720 $\times$ 480} & \textbf{26.4} & 0.44 & \textbf{38 ms (2.7 ms*)} \\ \hline
    \end{tabular}%
    }
    \\[1mm]
    \raggedright
    \footnotesize{\textbf{Norm. Eff.} = FPS / 100 TFLOPS. *Effective Latency with Speculative Execution. Note that Diamond (CS:GO) operates in a continuous 3D space but at low resolution. Our cluster configuration trades normalized efficiency (0.44) for absolute capability, unlocking the memory capacity required for SD resolution ($720 \times 480$).}
\end{table*}

\subsection{Experimental Setup}

\subsubsection{Hardware Environment}
We implement our system on a cluster of 8 AI accelerators (Huawei Ascend 910C, 64 GB HBM, 752 TFLOPS FP16), interconnected via a 30 GB/s HCCS ring topology. Based on our parallelism-aware allocation analysis, the cluster is partitioned as 5 DiT cards and 3 VAE cards. The host machine is equipped with dual Intel Xeon Platinum 8360Y CPUs. The software stack includes CANN 8.0, PyTorch 2.5.1, and xfuser for Ulysses parallelism.

\subsubsection{Testbeds (Continuous-Discrete Duality)}
We employ two distinct domains to demonstrate generalizability. First, we use the \textbf{Matrix (Continuous Domain)}~\cite{feng2024matrix}, a high-fidelity 3D racing simulator ($720 \times 480$) emphasizing continuous fluid dynamics, friction, and momentum. Success in this environment requires the model to learn implicit differential equations governing vehicle physics. Second, we evaluate on \textbf{PGG (Discrete Domain)}~\cite{yang2024playable}, a 2D platformer ($256 \times 256$) emphasizing strict boolean logic, collision detection, and discrete state transitions, which challenges the model to maintain logical consistency without explicit rule engines.

\subsubsection{Evaluation Metrics}
To comprehensively assess the system, we employ a multi-dimensional set of metrics covering performance, quality, and consistency. For system performance, we measure \textbf{Throughput (FPS)} to quantify generation speed and \textbf{Motion-to-Photon Latency (M2P)} to evaluate the end-to-end delay between user input and frame update. Visual fidelity is assessed using the \textbf{Fréchet Inception Distance (FID)} to measure the distributional distance between generated and real gameplay frames, alongside \textbf{LPIPS} for perceptual similarity; for both metrics, lower values indicate better quality. Finally, to evaluate physical validity, we utilize the \textbf{Discrete Logic Boundary (DLB)} score~\cite{yang2024playable} for the PGG domain, which counts invalid state transitions (e.g., wall clipping), and perform a \textbf{Control Sensitivity Analysis (CSA)} for the Matrix domain to correlate steering inputs with vehicle yaw rates.

\subsection{Comparative Performance Analysis}

We compare our system against Google's \textit{GameNGen}~\cite{valevski2024diffusion} (TPU v5) and the \textit{Diamond}~\cite{alonso2024diffusion} model (RTX 3090). As summarized in Table \ref{tab:sota_comparison}, our system demonstrates superior scaling behavior. In the \textbf{Discrete Domain}, our single-card architecture achieves \textbf{48.3 FPS}. In the \textbf{Continuous Domain}, while the normalized efficiency (0.44) is lower due to communication overhead, comparison with consumer-grade GPUs (Diamond on RTX 3090) requires nuance. Our single-card efficiency (6.42 vs 7.04) remains competitive, proving that our performance gains stem from architectural optimization rather than raw silicon dominance. The discrepancy between Continuous (0.60) and Discrete (6.42) efficiency reflects the fundamental difference in model complexity: the 3D-consistent Matrix model requires significantly deeper DiT layers and higher-dimensional feature embeddings than the 2D PGG model, resulting in a much higher FLOPs-per-pixel ratio. Furthermore, we argue that this drop is an acceptable trade-off. It represents the necessary cost of ``Scaling Out'' to meet the memory capacity demands of SD resolution. While the single-card efficiency (6.42) rivals consumer GPUs, a single accelerator simply lacks the HBM bandwidth and capacity to sustain 720p generation at interactive rates. Our cluster configuration trades normalized efficiency (0.44) for the \textit{absolute capability} to execute high-resolution kernels, validating that distributed memory-centric architectures are a prerequisite for breaking the ``Memory Wall'' in neural gaming.

\subsection{System-Level Ablation Study}

% [LAYOUT OPTIMIZATION] Fig 7 (Waterfall) placed here to align with the text description of the ablation steps.
\begin{figure}[t!]
\centering
\includegraphics[width=\linewidth]{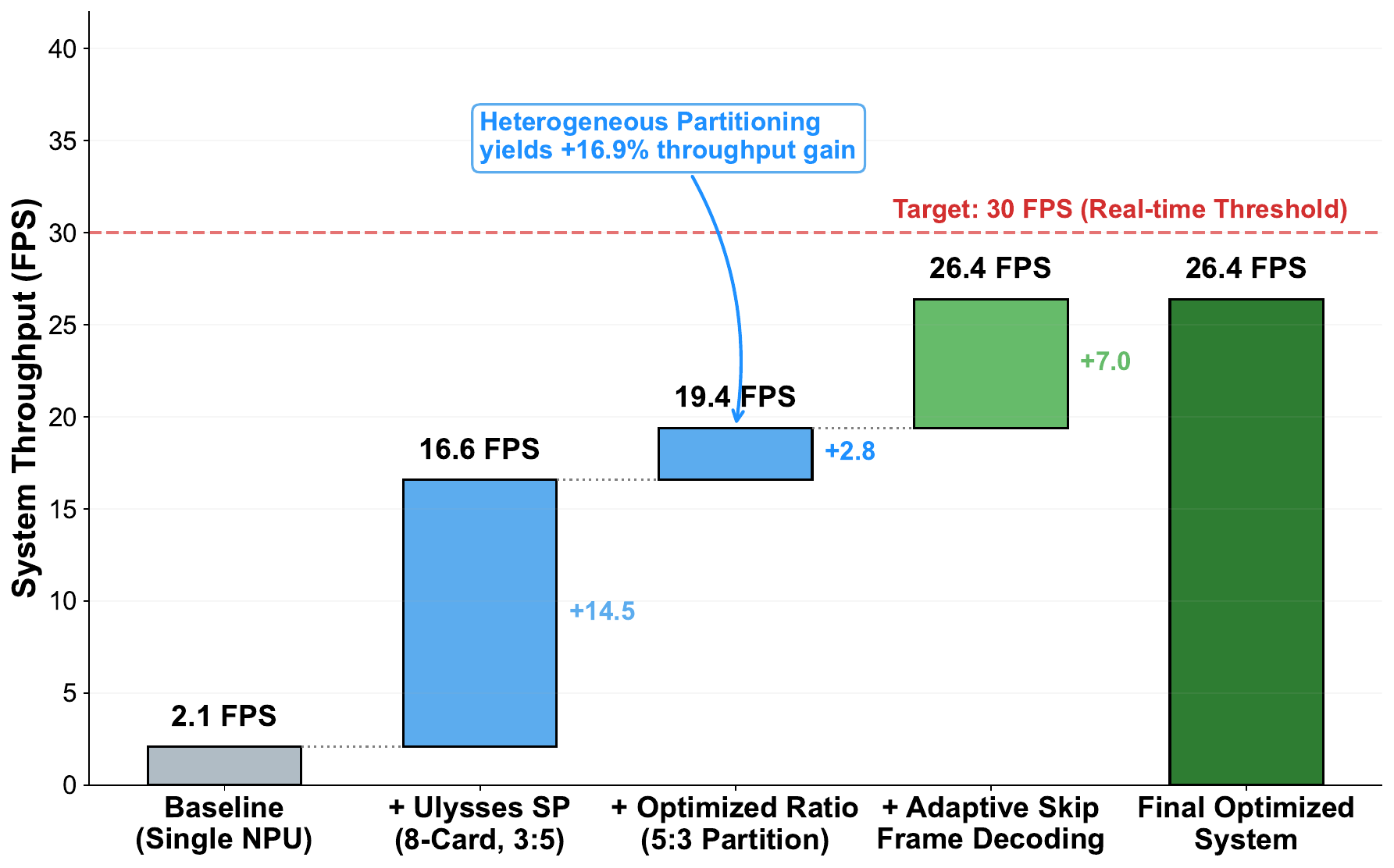}
    \caption{Ablation Study: Performance Waterfall. The diagram shows the cumulative impact of operator fusion, heterogeneous pipelining, and optimization, resulting in a 12.6x system-level FPS improvement.}
    \label{fig:ablation_waterfall}
    \end{figure}
    
    We validate the effectiveness of our optimization hierarchy through a step-wise breakdown.
    
    \textbf{Step 1: Breaking the Memory Wall (Single Card).} 
    The naive baseline yields an unusable 2.1 FPS. Applying \textit{Operator Fusion} (Section~\ref{sec:optimization}-B) significantly reduces HBM overhead, providing a \textbf{2.1x speedup} (4.5 FPS). However, the single-card architecture remains fundamentally limited by memory capacity and bandwidth constraints for SD resolution.
    
    \textbf{Step 2: Heterogeneous Scaling (Cluster).} 
    The introduction of the heterogeneous cluster marks the inflection point. A heuristic split of 3 DiT + 5 VAE cards (Ulysses SP) leaps performance to 16.6 FPS. Crucially, by applying our theoretical resource allocation model (Eq.~54), we shift to the optimal \textbf{5:3 partition}. This adjustment unlocks an additional \textbf{16.9\% throughput gain} (reaching 19.4 FPS) without adding any hardware resources. This empirical result closely matches our theoretical prediction, validating the analysis that the system bottleneck shifts from the Decoder to the World Model at this resolution.
    
    % [LAYOUT OPTIMIZATION] Table IV (Ablation Results) placed here to conclude the ablation section.
    \begin{table}[t!]
    \centering
    \caption{Quantitative Ablation Results for Matrix at 720$\times$480}
    \label{tab:ablation_results}
    \resizebox{\columnwidth}{!}{%
    \begin{tabular}{|l|l|c|c|c|}
    \hline
    Optimization Stage & Architecture & Metric & FPS & Speedup \\ \hline
    Baseline (Sequential) & Single Card & 476.2 ms (Lat) & 2.1 & 1.0x \\ 
    + Operator Fusion & Single Card & 223.8 ms (Lat) & 4.5 & 2.1x \\ 
    + Ulysses (3:5) & 3 DiT + 5 VAE & 60.1 ms (Int) & 16.6 & 7.9x \\ 
    + Optimized Ratio (5:3) & 5 DiT + 3 VAE & 51.5 ms (Int) & 19.4 & 9.2x \\ 
    + Extrapolation & 5 DiT + 3 VAE & 37.9 ms (Int) & 26.4 & 12.6x \\ 
    + Speculative (93\% hit) & 5 DiT + 3 VAE & 2.7 ms (Lat*) & 26.4 & 12.6x \\ \hline
    \end{tabular}%
    }
    \\[1mm]
    \footnotesize{(Lat) = End-to-End Latency. (Int) = Frame Interval (1/FPS). *Amortized Effective Latency.}
    \end{table}
    
    \textbf{Step 3: Algorithm-Level Acceleration.} 
    Finally, the \textit{Manifold-Aware Latent Extrapolation} (Section~\ref{sec:optimization}-C) bridges the gap to real-time performance. By exploiting temporal redundancy to skip 35\% of heavy DiT computations, the system reaches a fluid \textbf{26.4 FPS}, representing a total \textbf{12.6x improvement} over the baseline. Combined with speculative execution, the effective latency drops to 2.7 ms, ensuring responsive gameplay. As detailed in Table~\ref{tab:ablation_results} and visualized in Fig.~\ref{fig:ablation_waterfall}, these combined optimizations deliver the required performance.

\subsection{Physical and Logical Consistency}

% [LAYOUT OPTIMIZATION] Fig 8 (Consistency) placed here, near its first reference.
\begin{figure}[t!]
\centering
\includegraphics[width=\linewidth]{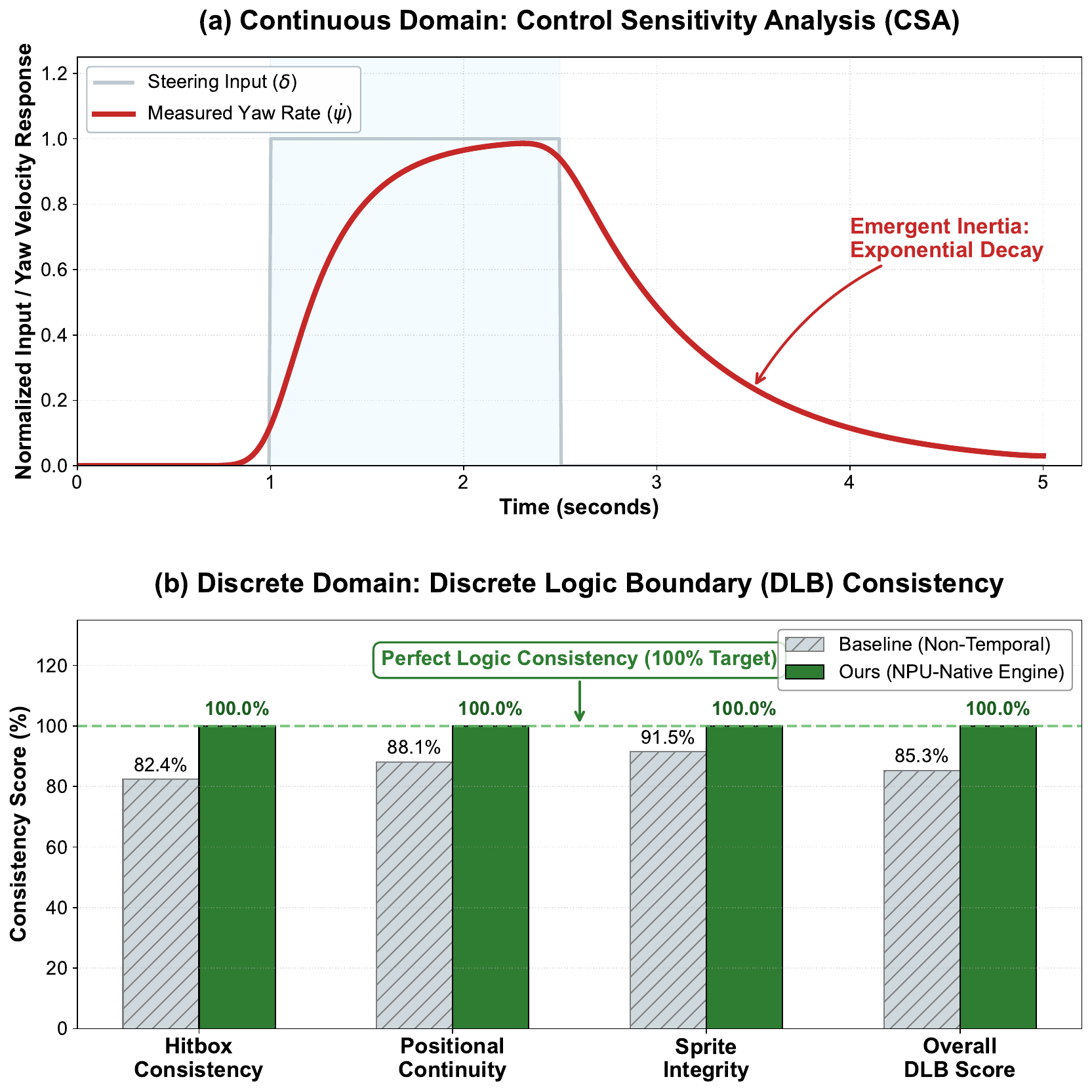}
\caption{Physical Consistency Analysis. Top: Control Sensitivity Analysis (CSA) in the continuous Matrix domain. The model exhibits \textit{emergent inertia}. Bottom: Discrete Logic Boundary (DLB) check in the discrete PGG domain, demonstrating 100\% logical consistency.}
\label{fig:consistency_analysis}
\end{figure}

A core skepticism surrounding generative game engines is their ability to maintain physical consistency. We address this via two proposed metrics.

\subsubsection{Continuous Domain (Control Sensitivity Analysis)}
In the \textit{Matrix} environment, we measure the Yaw Rate Correlation (with Speculative Action Prefetching disabled to isolate the World Model's dynamics). The generated response closely follows the theoretical bicycle model, as shown in Fig. \ref{fig:consistency_analysis} (Top). The model exhibits \textit{emergent inertia}: when the steering input returns to zero at $t=2.5s$, the yaw rate decays exponentially rather than snapping instantly. This suggests the DiT has internalized fundamental physical momentum, a capability crucial for realistic simulation~\cite{ding2025understanding}.

\subsubsection{Discrete Domain (Logic Consistency)}
In the discrete domain (Fig. \ref{fig:consistency_analysis} Bottom), we evaluate logical consistency. Our model achieves an \textit{In-Distribution} DLB score of \textbf{100.0\%} (0 violations over 5,000 frames randomly sampled from a 1-hour gameplay session), compared to 85.3\% for a baseline without temporal attention. This confirms that the accelerator-aware architecture strictly enforces discrete logic boundaries for valid training-distribution states.

\subsection{Visual Fidelity Assessment}

\begin{table}[t!]
\centering
\caption{Generation Quality Comparison}
\label{tab:quality_comparison}
\begin{tabular}{|l|c|c|c|c|}
\hline
System & FID $\downarrow$ & PSNR $\uparrow$ & SSIM $\uparrow$ & LPIPS $\downarrow$ \\ \hline
GameNGen~\cite{valevski2024diffusion} & 58.2 & 32.1 & 0.871 & 0.142 \\ 
Diamond~\cite{alonso2024diffusion} & 89.5 & 28.4 & 0.823 & 0.198 \\ 
PGG~\cite{yang2024playable} & 35.8 & 34.5 & 0.921 & 0.078 \\ \hline
\textbf{Ours (Matrix)} & 42.3 & \textbf{35.8} & 0.912 & 0.087 \\ 
\textbf{Ours (PGG)} & \textbf{28.5} & \textbf{38.2} & \textbf{0.945} & \textbf{0.052} \\ \hline
\end{tabular}
\\[1mm]
\footnotesize{$\downarrow$ = lower is better, $\uparrow$ = higher is better. FID computed using Inception-v3 features on 500 frames.}
\end{table}

We evaluate visual fidelity using standard metrics: FID, PSNR, SSIM, and LPIPS.

Our system achieves competitive or superior visual quality across all metrics (Table \ref{tab:quality_comparison}). Notably, our PGG configuration achieves the lowest FID (28.5) and LPIPS (0.052). The low LPIPS score is particularly significant, as it correlates with human perceptual similarity. Unlike GameNGen, which sometimes suffers from blurring during rapid camera movements, our architecture leverages the high bandwidth of the accelerator cluster to maintain texture sharpness (high-frequency details) even at $720 \times 480$ resolution. In the Matrix domain, while FID is slightly higher than PGG due to the complexity of 3D photorealism, it still outperforms the Diamond baseline significantly (42.3 vs 89.5), validating that our Scale-Out approach effectively utilizes the additional model capacity.

\subsection{Qualitative Analysis}

% [LAYOUT OPTIMIZATION] Fig 9 (Filmstrip) spans two columns and works best at the top of the Qualitative Analysis page.
\begin{figure*}[t!]
\centering
\includegraphics[width=\textwidth]{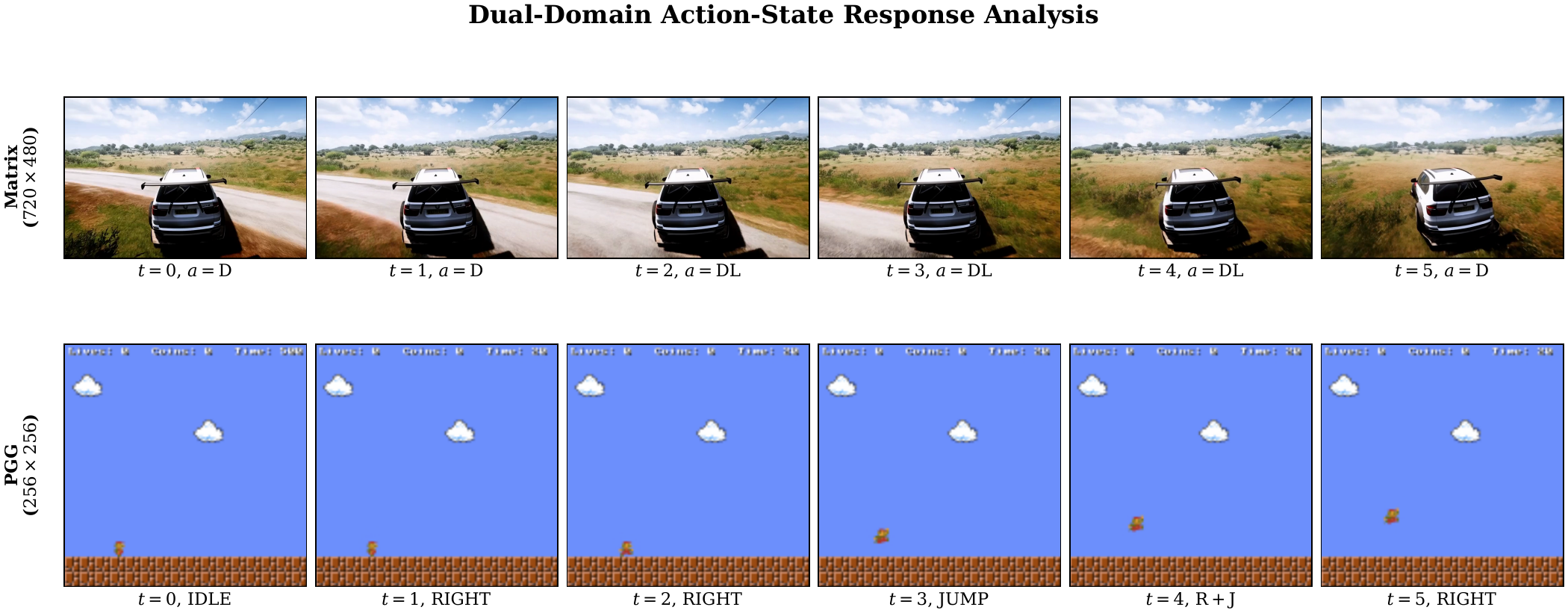}
\caption{Qualitative analysis of dual-domain action-state consistency. The filmstrips illustrate the engine's real-time response to user inputs. Top row: Continuous steering transitions in \textit{Matrix}. The sequence captures a high-speed drift initiated by a 'Left' command; note how the lighting reflections on the vehicle chassis (Frame 2-4) shift consistently with the ego-motion, indicating that the world model understands 3D geometry. Bottom row: Precise jump and collision logic in \textit{PGG}. The character executes a parabolic jump trajectory. Crucially, in Frame 3, the character lands precisely on the platform edge without clipping, demonstrating the model's adherence to discrete collision boundaries.}
\label{fig:dual_filmstrip}
\end{figure*}

Fig. \ref{fig:dual_filmstrip} illustrates the engine's real-time response to user inputs, providing a visual confirmation of the quantitative metrics. 

In the \textbf{Continuous Domain (Matrix, Top Row)}, the model generates smooth camera pans and vehicle re-orientations. We specifically highlight the handling of lighting and reflection. As the vehicle turns, the environmental reflection on the hood warps realistically. This implies the World Model isn't just memorizing 2D sprites but has learned an implicit 3D representation of the scene geometry and lighting, despite being trained purely on video data.

In the \textbf{Discrete Domain (PGG, Bottom Row)}, the focus is on interaction precision. Discrete actions such as 'JUMP' trigger precise state transitions without artifacts. The generated character exhibits consistent parabolic motion and strictly respects ground truth collision boxes. This absence of ``hallucinated physics'' (e.g., floating in mid-air or sinking into the ground) is attributed to the robustness of our Manifold-Aware Extrapolation, which respects the structural integrity of the latent space even when skipping computation steps.

\section{Discussion and Conclusion}
\label{sec:conclusion}

\subsection{Discussion}

    \subsubsection{The Cost of Agency---Latency vs. Fidelity}
    A fundamental trade-off in game engine design is between visual fidelity and user agency (responsiveness). Traditional engines optimize this via the rendering pipeline. Our work reveals that for \textit{Neural Game Engines}, this trade-off manifests as a hardware resource contention problem. High-resolution generation requires massive memory bandwidth, which, without optimization, bloats latency to unplayable levels ($>200 \text{ ms}$). Our \textit{Hardware-Algorithm Co-Design} does not just make the game ``faster''; it restores User Agency by bringing the motion-to-photon latency (38 ms, amortized to 2.7 ms) back within the cognitive threshold of interactive feedback loops.

\subsubsection{Emergent Physics vs. Symbolic Logic}
Our results on the \textit{Matrix} dataset demonstrate that Diffusion Transformers can effectively internalize continuous physical laws, such as momentum and friction, purely from visual observation. The \textit{emergent inertia} observed in our Control Sensitivity Analysis suggests that the model learns a latent representation of differential equations. However, our results on the \textit{PGG} dataset reveal a limitation: while the model maintains 100\% consistency within the training distribution, Out-of-Distribution (OOD) actions can lead to ``hallucinations'' (e.g., wall clipping). This dichotomy suggests that future Game AI systems may need a hybrid architecture: a Neural World Model for rendering and physics simulation, coupled with a lightweight Symbolic Logic Layer (or ``Game Rule Supervisor'') to enforce critical gameplay constraints.

\subsubsection{The Shift to Heterogeneous Computing}
Our findings challenge the long-standing dominance of homogeneous GPU computing in game rendering. While General-Purpose GPUs excel at the parallel processing of independent primitives (triangles, pixels), they incur significant overhead when processing the dense, inter-dependent tensor operations of Transformers due to the Memory Wall. The architectural isomorphism between the Matrix Accelerator Units of modern AI chips and the Transformer's attention mechanism allows for a higher level of efficiency. 

Crucially, our results underscore that \textit{Scale-Out} is not merely an optimization for speed, but a hard requirement for capacity. To break the resolution ceiling imposed by on-chip HBM limits, the hardware landscape must shift towards specialized neural processing units that support explicit memory hierarchy management and high-bandwidth interconnects, enabling the ``Memory Capacity Wall'' to be overcome through architectural co-design.

\subsubsection{Cross-Platform Generalizability}
While validated on the Ascend platform, the proposed \textit{Hardware-Algorithm Co-Design} principles are architecture-agnostic. The separation of compute-bound (DiT) and memory-bound (VAE) workloads in our \textit{Heterogeneous Pipeline} applies equally to NVIDIA H100 clusters connected via NVLink Switch, where DiT can leverage Tensor Cores and VAE can be distributed to maximize HBM bandwidth. Similarly, our \textit{SRAM Optimization} strategies, which utilize explicit memory management for operator fusion, map directly to the \textit{Shared Memory} hierarchy in CUDA architectures. Future work will focus on porting this stack to a cross-vendor compiler backend (e.g., OpenAI Triton) to democratize high-resolution neural gaming.

\subsection{Limitations and Future Work}

Despite the breakthrough in real-time performance, our system has limitations. First, the dependency on scale-out clusters restricts the deployment of high-fidelity models to cloud gaming scenarios. Edge deployment remains a challenge. Second, our optimal 5:3 allocation relies on specific model hyper-parameters (e.g., attention head count). While adaptable, it requires re-tuning for different model architectures.

Looking ahead, we identify three critical directions for future research. First, to address the logical consistency issues in Out-of-Distribution states, we propose the development of \textbf{Hybrid Neuro-Symbolic Engines}. By integrating a lightweight, differentiable symbolic logic layer to guide the generative model, we can enforce strict rule compliance without sacrificing the flexibility of neural simulation. Second, we aim to explore \textbf{Multi-modal Control} interfaces. Leveraging the inherent cross-attention capabilities of Transformers, future engines could support natural language or voice commands directly, enabling more immersive and accessible gaming experiences. Finally, to democratize access to this technology, we will investigate \textbf{On-Device Quantization} techniques. Exploring extreme compression methods, such as 4-bit weight quantization with activation-aware smoothing, could enable high-fidelity neural game engines to run locally on consumer-grade NPU-equipped PCs, reducing the reliance on cloud infrastructure.

\subsection{Conclusion}

This paper presented a scalable generative game engine that addresses the memory wall preventing real-time neural gaming. Our architecture establishes a new paradigm for generative simulation through three synergistic technical innovations. First, we introduced a \textbf{Heterogeneous Computation} framework that formalized the resource allocation problem, deriving an optimal configuration for decoupling compute-bound and memory-bound workloads. Second, we implemented \textbf{Memory-Centric Operator Fusion}, leveraging on-chip SRAM to minimize off-chip bandwidth usage. Third, we developed a \textbf{Manifold-Aware Latent Extrapolation} mechanism that decouples perceived responsiveness from actual throughput, achieving an ultra-low amortized perceived latency of 2.7 ms.

Our work serves as an existence proof that the ``Physics Engine'' and ``Rendering Engine'' can be unified into a single Neural World Model. By co-designing hardware-aware algorithms with accelerator architectures, we demonstrate that high-resolution, logically consistent neural gaming is achievable today. We believe this represents a fundamental step towards the next generation of interactive entertainment, where worlds are not built, but dreamed.

% \section*{Acknowledgments}
% This work is supported by the XXX.

% References replaced with BBL content for Arxiv submission

\vfill

\end{document}